\documentclass{article}
\usepackage{graphicx} 
\usepackage{subcaption} 
\usepackage{amsmath}
\usepackage{amssymb}
\usepackage{bbm}
\usepackage{hyperref}

\usepackage{algorithm}
\usepackage{algpseudocode}
\usepackage{url}
\usepackage{iclr2024/iclr2024_conference,times}
\algrenewcommand\algorithmicrequire{\textbf{Input:}}
\algrenewcommand\algorithmicensure{\textbf{Output:}}
\usepackage{xcolor}

\usepackage{amsmath,amsfonts,bm}

\def\eqref#1{equation~\ref{#1}}
\def\Eqref#1{Equation~\ref{#1}}

\def\1{\bm{1}}

\DeclareMathAlphabet{\mathsfit}{\encodingdefault}{\sfdefault}{m}{sl}
\SetMathAlphabet{\mathsfit}{bold}{\encodingdefault}{\sfdefault}{bx}{n}

\title{Latent-EnSF: A Latent Ensemble Score Filter for High-Dimensional Data Assimilation with Sparse Observation Data}
\author{Phillip Si, Peng Chen}
\date{July 2024}

\begin{document}

\maketitle

\begin{abstract}
Accurate modeling and prediction of complex physical systems often rely on data assimilation techniques to correct errors inherent in model simulations. Traditional methods like the Ensemble Kalman Filter (EnKF) and its variants as well as the recently developed Ensemble Score Filters (EnSF) face significant challenges when dealing with high-dimensional and nonlinear Bayesian filtering problems with sparse observations, which are ubiquitous in real-world applications. In this paper, we propose a novel data assimilation method, Latent-EnSF, which leverages EnSF with efficient and consistent latent representations of the full states and sparse observations to address the joint challenges of high dimensionlity in states and high sparsity in observations for nonlinear Bayesian filtering. We introduce a coupled 
Variational Autoencoder (VAE) with two encoders to encode the full states and sparse observations in a consistent way guaranteed by a latent distribution matching and regularization as well as a consistent state reconstruction. With comparison to several methods, we demonstrate the higher accuracy, faster convergence, and higher efficiency of Latent-EnSF for two challenging applications with complex models in shallow water wave propagation and medium-range weather forecasting, for highly sparse observations in both space and time.

\end{abstract}

\section{Introduction}

Many complex physical systems are traditionally modeled by partial differential equations (PDEs). First-principle PDE-based modeling and simulation have been demonstrated to be powerful in making predictions of complex systems in various scientific and engineering fields. However, in practical applications, significant discrepancy of the simulation-based prediction and the reality may arise from different sources, e.g., model inadequacy, uncertainties in model parameters, boundary and/or initial conditions, exteral forcing/loading terms, numerical approximation errors, etc. Meanwhile, data-driven machine learning (ML)  models have been significantly developed and employed to make system predictions in the last few years, which have been shown to be extremely efficient in the case of large dynamical systems such as weather forecasting, a problem that requires the supercomputer of hours to run for the European Centre for Medium-Range Weather Forecasts (ECMWF), but one that can be solved by neural network surrogates such as FourCastNet \citep{pathak2022fourcastnetglobaldatadrivenhighresolution}
in a matter of seconds. These properties also make them particularly suitable for ensemble forecasting and uncertainty quantification. However, these models are also often subject to the key limitation of divergence from reality in the long run, resulting in significant accumulated errors for long-term predictions because of various uncertainties and the autoregressive structure of the ML models.

To mitigate the discrepancy and accumulated errors and make more accurate predictions, data assimilation plays an essential role in  
incorporating additional observation data of the reality into the current PDE/ML-based prediction models. By adapting the PDE/ML state to match the observation data, the accuracy of future state predictions can be markedly improved. 
Data assimilation \citep{sanzalonso2023inverseproblemsdataassimilation, doi:10.1137/1.9781611974546} approaches such as Kalman filter (KF) \citep{kalman1960new}, particle filters \citep{knsh_2013}, and their variants have been widely applied to problems such as weather prediction, geophysical modeling, robotics, and many more areas.

The basis of many current data assimilation methods used in practice is the KF \citep{kalmanfilter}. It parameterizes its state with the mean and the covariance, and presents the optimal solution to the data assimilation problem given that certain linearity properties of the transition and observation function are fulfilled. However, in practice, KFs end up being computationally expensive for high-dimensional states and biased for nonlinear problems.
An alternative approach is particle filter, which represents the density by an ensemble of particles. They relax the assumption of distributional form, e.g., Gaussian distribution, made for distribution-based approaches such as the KF. The EnKF \citep{kalman1960new} is a particular particle filter that originates from the KF. Instead of keeping the covariance matrix, it computes the sample covariance of the ensemble of particles. Due to its robust performance, it has become a widely used method for applications today. 
However, most particle filters, including EnKF, suffer from the curse of dimensionality of requiring an exponentially large number of particles to accurately describe the distribution in high dimensions.
To address this challenge, various extensions to EnKF have been proposed, such as the Localized Ensemble Transform Kalman Filter (LETKF) \citep{HUNT2007112}, where covariance localization techniques as well as advancements from the Ensemble Transform Kalman Filter \citep{AdaptiveSamplingwiththeEnsembleTransformKalmanFilterPartITheoreticalAspects} are applied.

Nonetheless, a few major roadblocks hinder efficient data assimilation for high dimensional systems. Particularly, standard data assimilation approaches have quite limited capabilities in high-dimensional environments, resorting to assumptions such as localization in order to make them more applicable. High dimensional systems and complex models can also limit the capability and efficiency of models such as 4D-Var \citep{rabier20034dvar}, which is extremely inefficient since it requires propagating through the physical model many times. A recent approach named DiffDA \citep{huang2024diffdadiffusionmodelweatherscale} uses diffusion models and masked interpolations of the weather data to assimilate sparse observations, but the approach falls behind pure interpolation after a few steps. \cite{bao2024ensemblescorefiltertracking} proposed a novel method called Ensemble Score Filter (EnSF) which has achieved great results on high-dimensional data assimilation problems by leveraging the power of score-based diffusion models to sample from the posterior distribution efficiently. However, sparse observations prove to be very challenging for EnSF, limiting its applicability to real-world scenarios where observation data are limited. Meanwhile, approaches such as Generalized Latent Assimilation (GLA) \citep{cheng2022generalised} attempt to use dimension-reduction techniques to conduct data assimilation in latent space, but are still subject to the constraints of the standard EnKF for complex high-dimensional systems.

\begin{figure}[h]
    \centering
    \includegraphics[width=\textwidth]{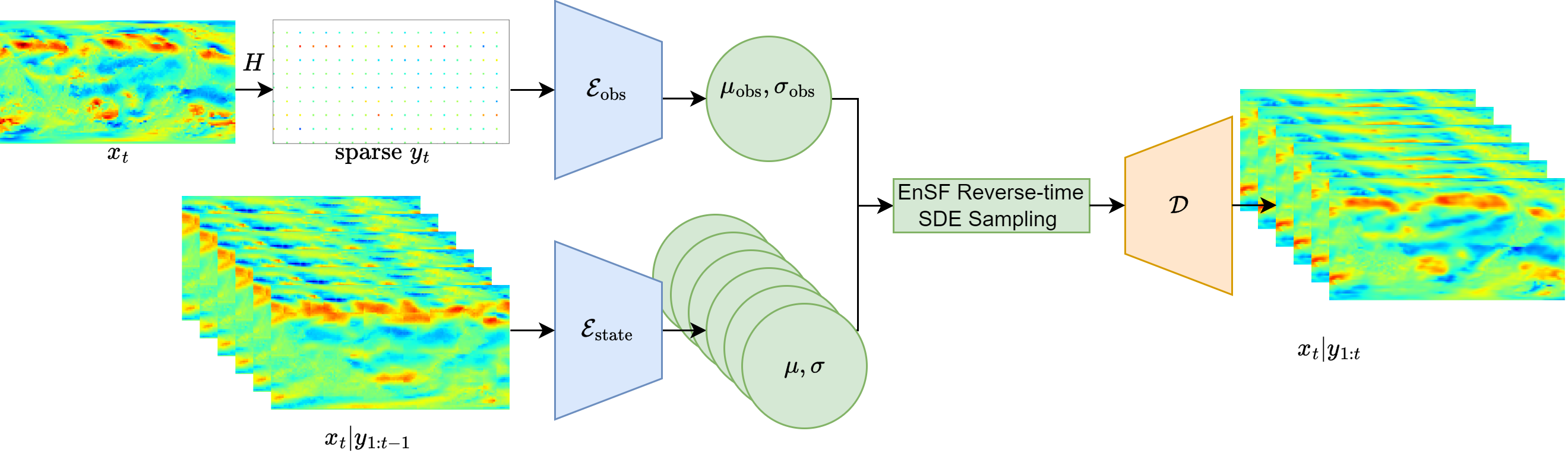}
    \caption{Flow of the Latent-EnSF. An ensemble of prior states $x_{t}|y_{1:t-1}$ is assimilated with sparse observation $y_{t}$ in the latent autoencoder space to obtain samples $x_{t}|y_{1:t}$ from the posterior.}
    \label{fig:ga_ensf}
\end{figure}

In this work, we propose an efficient data assimilation method called Latent-EnSF that builds on Ensemble Score Filters (EnSF) \citep{bao2024ensemblescorefiltertracking} to address the joint challenges of high dimensionality and data sparsity. See Figure \ref{fig:ga_ensf} for a workflow of the Latent-EnSF. We propose to construct coupled Variational Autoencoders (VAE) \citep{kingma2014autoencoding} to encode the sparse observations into a latent space in which we conduct data assimilation by EnSF using diffusion-based sampling. The ensemble of samples of the assimilated latent state is then decoded to an ensemble of samples of the full state. Importantly, our paper addresses the key weakness of the EnSF when dealing with sparse observations, and retains its ability to efficiently assimilate observations for high-dimensional nonlinear systems. We demonstrate this performance for both PDE and ML based prediction models of some challenging examples, experimenting with data that is sparse in both space and time.

We will begin by formally introducing data assimilation problems with a review of the EnSF in Section \ref{sec:DA}, follow up with how we integrate this with coupled variational autoencoders in Section \ref{sec:LatentEnSF} to create our Latent-EnSF, and finally conclude with experiments in Section \ref{sec:experiments}. 

\section{Dynamical Systems and Data Assimilation}
\label{sec:DA}
To model a dynamical physical system, we denote the state at a certain time $t$ as $x_t \in \mathbb{R}^d$, where $d$ is typically very high for complex models. For a system with discrete time steps $t = 1, 2, \dots$, given initial condition $x_0$, we then model the evolution of the state from the time $t-1$ to time $t$ as
\begin{equation} \label{eq:transition}
x_{t} = M(x_{t-1}, \varepsilon_{t}),
\end{equation}
where $\varepsilon_{t}$ is a process noise term coming from a known distribution; this term accounts for hidden interactions and numerical errors. The corresponding observation $y_t \in \mathbb{R}^m$ at time $t$ is given by 
\begin{equation} \label{eq:observation}
y_t = H(x_t) + \gamma_t,
\end{equation}
where $\gamma_t$ (also coming from a known distribution, typically Gaussian) represents the observation noise because of instrumental inaccuracies, etc. Both the model map $M: \mathbb{R}^d \times \mathbb{R}^d \to \mathbb{R}^d$ and the observation map $H: \mathbb{R}^d \to \mathbb{R}^m$ can be nonlinear.

Now, the model of the dynamical system in \Eqref{eq:transition} may lead to inaccurate representation of the ground truth with increasing discrepancy in time because of, e.g., model inadequacy or input uncertainty. The goal of data assimilation is to assimilate the observation data 
of the true state 
to the model of the dynamical system and recover an accurate representation of the true state.
Note that in practical applications the observation data 
can be sparsely distributed in both space and time, 
which makes the recovery of the true state very challenging, especially for high dimensional problems with the state dimension $d \gg 1$, and the observation dimension $m$ much smaller than $d$.   

\subsection{Bayesian Filtering}
\label{section:bayesianfiltering}

From a Bayesian filtering perspective \citep{bayesianfiltering}, a data assimilation problem at each time $t$ can be divided into two steps: a prediction step that advances the dynamical system and an update step that assimilates the data.
Assuming that the posterior density of the state $x_{t-1}$ given observation data $y_{1:t-1} = (y_1, \dots, y_{t-1})$, denoted as $P(x_{t-1} | y_{1:t-1})$, is available at time $t-1$, with $P(x_0|y_{1:0}) = P(x_0)$ given for time $t-1 = 0$, then the prediction step provides the density of $x_{t}$ as
\begin{equation} \label{eq:prediction}
\textbf{Prediction: } P(x_{t} | y_{1:t-1}) = \int P(x_{t} | x_{t-1}) P(x_{t-1} | y_{1:t-1}) dx_{t-1},
\end{equation}
where $P(x_{t} | x_{t-1})$ represents the transition probability governed by the dynamical system in Equation \ref{eq:transition}. Note that $x_t$ depends on $y_{1:t-1}$ only through $x_{t-1}$.
Let $P(y_{t} | x_{t})$ denote the likelihood function of the data $y_{t}$ given state $x_{t}$ from Equation \ref{eq:observation}. Then, the update step provides the updated posterior density $P(x_{t} | y_{1:t})$ of the state $x_{t}$ given the new observation data $y_{t}$ by Bayes' rule as 
\begin{equation} \label{eq:bayes_filtering}
\textbf{Update: } P(x_{t} | y_{1:t}) = \frac{P(y_{t} | x_{t}) P(x_{t} | y_{1:t-1})}{P(y_{t} | y_{1:t-1})},
\end{equation}
where the normalization constant or model evidence term $P(y_{t} | y_{1:t-1})$ is given by $P(y_{t} | y_{1:t-1}) = \int P(y_{t} | x_{t}) P(x_{t} | y_{1:t-1}) dx_{t}$, which is typically intractable to compute.

\subsection{Diffusion Models and the Ensemble Score Filter}
\label{sec:DMEnSF}

A recent development for a new particle filter is the EnSF, where the ensemble of samples can be drawn from the posterior distribution through a diffusion process. It builds on the recent advancement in score-based generative modeling \citep{ho2020denoising} in generating high-fidelity samples from a distribution by first computing a score function $\nabla_x P(x)$ and then sampling with it by solving stochastic differential equations (SDE) \citep{song2021scorebasedgenerativemodelingstochastic}. This circumvents knowing the true density as it only needs to generate samples from the posterior distribution.
Notably, the diffusion model takes the form of a noisy forward stochastic differential equation process
\begin{equation} \label{eq:forward_sde}
dx = f(x, \tau)d\tau + g(\tau)dw,
\end{equation}
which transforms data from an arbitrary distribution to an isotropic Gaussian for $\tau \in \mathcal{T} = [0, T]$. Here, $f(x, \tau)$ is a drift term, $g(\tau)$ is a diffusion term, and $w$ is a $d$-dimensional Wiener process. It has the corresponding reverse-time SDE that runs backward in time from $\tau = T$ to $0$ as
\begin{equation} \label{eq:reverse_sde}
dx = [f(x, \tau) -  g^2(\tau)\nabla_x \log P_{\tau}(x)]d\tau + g(\tau) d\bar{w},
\end{equation}
where $\bar{w}$ is another Wiener process independent of $w$. This reverse-time SDE can be solved to sample from the distribution $P(x)$ if the score function $\nabla_x \log P(x_{\tau})$ is known, e.g., by the Euler--Maruyama scheme at the discrete time steps $0=\tau_0<\tau_2<...<\tau_k = T$. 

In the following part, we use $x_{t, \tau}$ to indicate the state at physical time $t$ and diffusion time $\tau$.
The EnSF method \citep{bao2024ensemblescorefiltertracking} uses $T = 1$ and the following drift and diffusion terms 
\begin{equation}\label{eq:fg}
    f(x_{t, \tau}, \tau) = \frac{d \log \alpha_{\tau}}{d \tau} x_{t, \tau} \text{ and } g^2(\tau) = \frac{d \beta_{\tau}^2}{d \tau} - 2\frac{d \log \alpha_{\tau}}{d \tau}\beta_{\tau}^2, 
\end{equation}
with $\alpha_{\tau} = 1 - \tau (1- \epsilon_\alpha)$ and $\beta^2_{\tau} = \epsilon_\beta + \tau(1-\epsilon_\beta)$ for two small positive hyperparameters $\epsilon_\alpha$ and $\epsilon_\beta$ which aid in avoiding the collapse of the generated samples.  
This choice leads to the distribution $x_{t, \tau} \sim N(\alpha_\tau x_{t, 0}, \beta^2_\tau I)$ conditioned on $x_{t, 0} = x_t$, which results in the prior score function 
\begin{equation}\label{eq:prior_score}
\begin{split}
    \nabla_x \log P(x_{t, \tau}|y_{1:t-1}) & = \nabla_x \log \int P(x_{t, \tau}| x_{t}) P(x_{t}|y_{1:t-1}) d x_{t} \\
    & = \int -\frac{x_{t, \tau} - \alpha_\tau x_{t}}{\beta_{\tau}^2} \omega(x_{t, \tau}, x_{t}) P(x_{t} | y_{1:t-1}) dx_{t}, 
\end{split}
\end{equation}
where the gradient $\nabla_x$ is taken with respect to $x_{t,\tau}$, and the weight $\omega(x_{t, \tau}, x_{t})$ is given by 
\begin{equation} \label{eq:weight}
    \omega(x_{t, \tau}, x_{t}) = \frac{P(x_{t, \tau}|x_{t})}{\int P(x_{t, \tau}|x_{t}')P(x_{t}' | y_{1:t-1})dx_{t}'}.
\end{equation}
Both the score function in \Eqref{eq:prior_score} and the weight in \Eqref{eq:weight} can be evaluated by sample average approximation with random samples from the distribution $P(x_{t} | y_{1:t-1})$ obtained in the prediction step in \Eqref{eq:prediction}. In the update step in \Eqref{eq:bayes_filtering}, the posterior score function in the EnSF method \citep{bao2024ensemblescorefiltertracking} is formulated as 
\begin{equation} \label{eq:scorefunction}
\nabla_x \log P(x_{t, \tau} | y_{1:t}) = \nabla_x \log P(x_{t, \tau} | y_{1:t-1}) + h(\tau) \nabla_x \log P(y_{t}|x_{t, \tau}),
\end{equation}
with the prior score function in the first term given in \Eqref{eq:prior_score}, the likelihood function $P(y_{t}|x_{t, \tau})$ in the second term given explicitly from the observation map in \Eqref{eq:observation}, and a damping function $h(\tau)$ that is monotonically decreasing with $h(0) = 1$ and $h(1) = 0$, e.g., $h(\tau) = 1 - \tau$. Note that the posterior score function is consistent with the Bayes' rule for the posterior in \Eqref{eq:bayes_filtering} at $\tau = 0$. 





To this end, a brief description of one step of the EnSF algorithm is presented in Algorithm \ref{alg:ensf}.
\begin{algorithm}
\caption{One Step of EnSF}\label{alg:ensf}
\begin{algorithmic}
\Require Ensemble of the states $\{x_{t-1}\}$ from distribution $P(x_{t-1} | y_{1:t-1})$ and the observation $y_{t}$.
\Ensure Ensemble of the states $\{x_{t}\}$ from the posterior distribution $P(x_{t} | y_{1:t})$.
\State Run the (stochastic) dynamical system model in \Eqref{eq:transition} from $\{x_{t-1}\}$ to obtain samples $\{x_t\}$. 
\For{$\tau = \tau_k,...,\tau_0$}
\State Estimate the prior score $\nabla_x \log P(x_{t, \tau} | y_{1:t-1})$ using \Eqref{eq:prior_score} and \ref{eq:weight}.
\State Estimate the posterior score $\nabla_x \log P(x_{t, \tau} | y_{1:t})$ using \Eqref{eq:scorefunction}.
\State Solve the reverse-time SDE in \Eqref{eq:reverse_sde} to generate the ensemble $\{x_t\}$. 
\EndFor
\end{algorithmic}
\end{algorithm}

There are a number of advantages that EnSF provides over EnKF, particularly in high dimensions, where a much larger number of ensemble members are typically required for EnKF to have a good estimate of the approximate Gaussian density, while EnSF leverages the explicit likelihood function and the diffusion process to generate samples. In addition, EnSF does not assume approximate linearity of the dynamical system, making it highly applicable to high-dimensional and nonlinear filtering problems, e.g., a million-dimensional Lorenz-96 system \citep{bao2024ensemblescorefiltertracking} and a surface quasigeostrophic model \citep{bao2024nonlinearensemblefilteringdiffusion}. However, the key limitation of EnSF to filtering with sparse observation arises also from the explicit use of the likelihood function, as explained below.

\subsection{Sparse Observations} \label{section:ensf_sparsity}

\begin{figure}[h]
    \centering
    \includegraphics[width=0.8\textwidth]{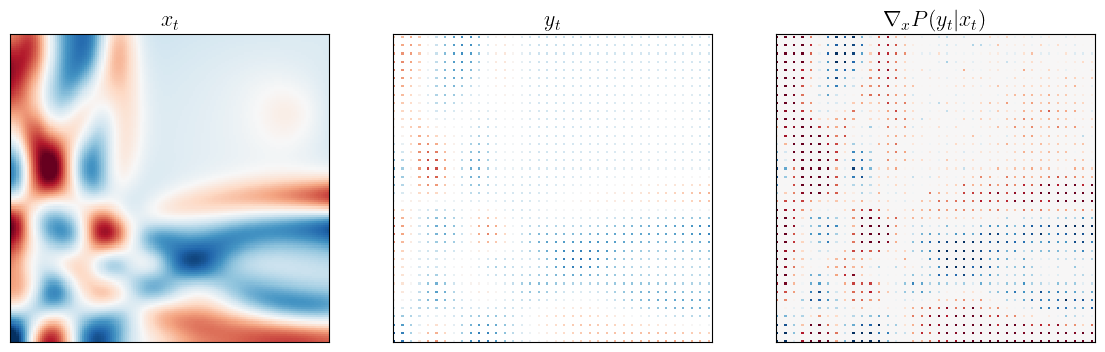}
    \caption{The gradient of the log-likelihood function $\nabla_x P(y_t|x_t)$ (right) vanishes at the points where the sparse observation data $y_t$ (middle) do not have any information of the state $x_t$ (left).}
    \label{fig:score_function}
\end{figure}

A key limitation of EnSF in high-dimensional nonlinear filtering problems occurs when the observation data are sparse, which is the case for most practical applications.
For example, suppose that the observation map $H(x_t) = x_t[S]$, which only make observations of $x_t$ in the dimensions from the subset $S \subset \{1,...,d\}$, where $S$ may have a much smaller cardinality $|S| \ll d$ than the full dimension. In this case, the gradient of the log-likelihood $\nabla_x \log P(y_{t}|x_{t})$ vanishes in the dimensions in $\{1,...,d\}/S$, as shown in Figure \ref{fig:score_function}, where we calculate $\nabla_x P(y_t|x_t)$ for one of our experiments in Section \ref{section:shallow_water} with 1/16 of the total dimensions used for the observations. Note that, the gradient is nonzero only at the points which are observed as seen in Figure \ref{fig:score_function}. Though the ramifications are minimal for sufficiently dense observations, the vanishing gradients for higher sparsity scenarios significantly limits the effectiveness of the standard EnSF, which we demonstrate empirically in section \ref{section:ensf_sparsity_experiments}. 

To mitigate this weakness of the EnSF, 
rather than conducting data assimilation in the full state space with sparse observations, we propose to transform the data assimilation into a learned latent space with sufficiently preserved information and significantly reduced dimension.


\section{Latent ensemble score filter}
\label{sec:LatentEnSF}

To address the key limitation of EnSF in the case of sparse observations due to the vanishing gradient of the log-likelihood function, we propose a latent representation of the sparse observations by a variational autoencoder (VAE) and match this latent representation to the encoded full state for the sake of consistent data assimilation in the latent space, for which we employ EnSF with the additional advantage of VAE regularization of the latent variables.

The advantages are twofold: it allows us to not only make use of the benefits of standard latent score-based generative models \citep{NEURIPS2021_5dca4c6b, Rombach_2022_CVPR} which offer optimized sampling speed and expressivity, but also create a expressive map from a sparse observation space to a full-dimensional latent space. Additionally, the successful application of latent-diffusion models to videos \citep{Blattmann_2023_CVPR} offers a potential application of exploring latent-assimilation coupled with latent dynamics.

\subsection{Compression by variational autoencoder}

The Variational Autoencoder (VAE) \citep{kingma2014autoencoding} provides a compressed representation of a high-dimensional distribution by means of bottlenecking. 
Note that since this is a family of models, it admits many specific instances such as VQ-VAE \citep{oord2018neural}, InfoVAE \citep{zhao2018infovae}, etc., which can promote desirable properties such as independence and adaptable priors in the latent space. VAEs have two components, an encoder $\mathcal{E}$ and a decoder $\mathcal{D}$. The encoder casts the original state $x \in \mathbb{R}^d$ into a low dimensional representation $\mathcal{E}(x) \in \mathbb{R}^{2r}$ such that $2r \ll d$. Then, the model splits $\mathcal{E}(x)$ into mean $\mu \in{\mathbb{R}^{r}}$ and variance $\sigma^2 \in{\mathbb{R}^{r}}$, and samples from the normal distribution $N(\mu, \Sigma)$ with $\Sigma = \text{diag}(\sigma_1^2, \dots, \sigma_r^2)$ by applying a reparameterization trick $z = \mu + \sigma \cdot \epsilon$, where $\epsilon \in N(0, I_r)$. The decoder then constructs an approximation of the state as $\tilde{x} = \mathcal{D}(z)$.

VAEs are trained by optimizing for the evidence lower bound (ELBO), which is composed of a reconstruction term in the state space (e.g. the MSE $||x - \tilde{x}||_2^2$) and a Kullback–Leibler divergence (KLD) regularization term in the latent space, 
which is given by 
\begin{equation} \label{eq:KLD}
\text{KLD}(\mu, \Sigma) = \sum_{i = 1}^r -\log(\sigma_i) + \frac{\sigma_i^2+\mu_i^2}{2} - \frac{1}{2},
\end{equation} 
which skews $N(\mu, \Sigma)$ to be close to $N(0, I_r)$.



\subsection{Coupled VAEs with Latent-Space Matching}
\label{sec:LRSO}

\begin{figure}[h]
    \centering
    \includegraphics[width=0.8\textwidth]{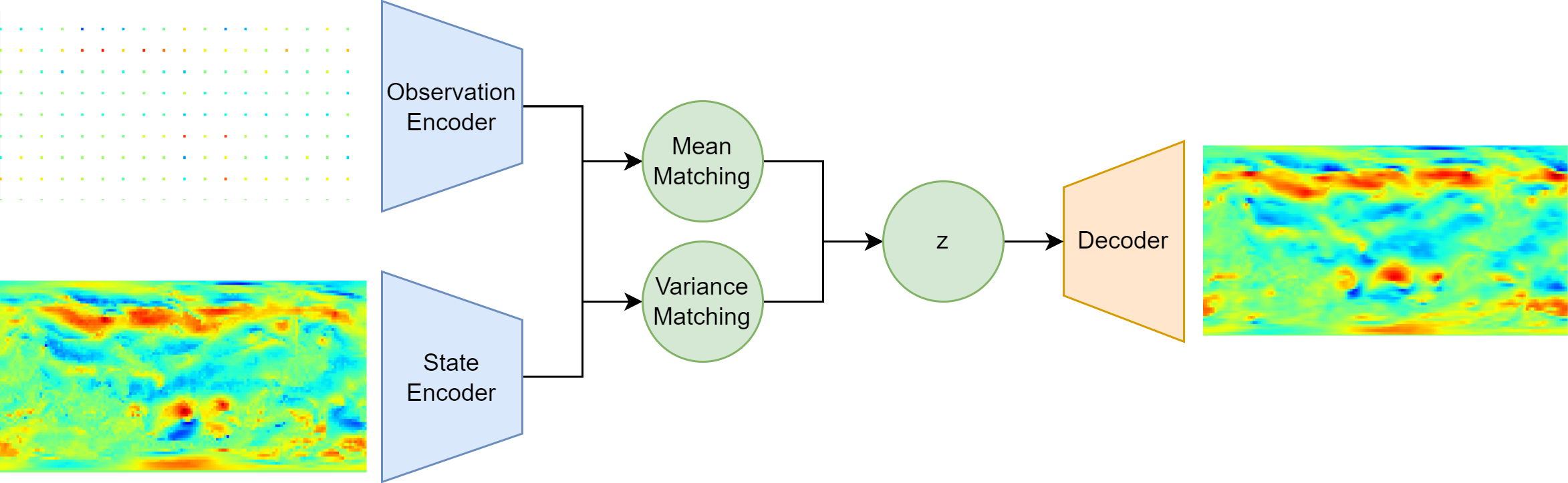}
    \caption{A coupled VAE for consistent latent representations of sparse observations and full states. }
    \label{fig:vae_obsmatching}
\end{figure}

We use VAEs to compress both the sparse observations and the full states to the latent space. To achieve consistent latent representations of sparse observations and full states, we couple the VAEs in the latent space as shown in Figure \ref{fig:vae_obsmatching}. 


Specifically, we formulate a coupled VAE with two encoders, one encoder $\mathcal{E}: \mathbb{R}^d \rightarrow \mathbb{R}^{2r}$ that encodes the full state $x_t$ to a latent variable $z_t$ with distribution $N(\mu_t, \Sigma_t)$, where $(\mu_t, \sigma^2_t) = \mathcal{E}(x_t)$, and $z_t = \mu_t + \sigma_t \cdot \epsilon_t$, and the other encoder $\mathcal{E}_{\text{obs}}: \mathbb{R}^m \rightarrow \mathbb{R}^{2r}$ that encodes the sparse observation $y_t$ to another latent variable $z_t^{\text{obs}}$ with distribution $N(\mu^{\text{obs}}_t, \Sigma^{\text{obs}}_t)$, where $(\mu^{\text{obs}}_t, (\sigma_t^{\text{obs}})^2) = \mathcal{E}_{\text{obs}}(y_t)$ and $z_t^{\text{obs}} = \mu^{\text{obs}}_t + \sigma^{\text{obs}}_t \cdot \epsilon_t$. 
To assimilate the encoded latent data to the encoded latent state, we seek to match the two latent representations for their means and variances. To build a consistent reconstruction of the full state from the latent representation, we use the same decoder $\mathcal{D}: \mathbb{R}^r \to \mathbb{R}^d$. To this end, we formulate the following loss function to account for all three contributions: 
\begin{align}
    \ell_{t}(\theta) &= ||x_t - \mathcal{D}(z_t)||_2^2 + ||x_t - \mathcal{D}(z^{\text{obs}}_t)||_2^2 & \text{ (Reconstruction Term)}\\        
    &+ \text{KLD}(\mu_t, \Sigma_t) + \text{KLD}(\mu^{\text{obs}}_t, \Sigma^{\text{obs}}_t) & \text{ (Regularization Term)}\\
    &+ ||\mu_t - \mu^{\text{obs}}_t||_2^2 + ||\Sigma_t - \Sigma^{\text{obs}}_t||_2^2 & \text{ (Latent Matching Term)},
\end{align}
where $\theta$ represents the parameters of the encoders and decoder, for which we use convolutional neural networks (CNNs) in this work. Then for the training of the parameter $\theta$ in the coupled VAE, we minimize an empirical loss function as a sum of the above loss function over the training trajectories of the dynamical system, and over all the time steps $t$ in each trajectory. 

Note that both of the latent representations use the same decoder for reconstruction of the full state. By training a coupled VAE which minimizes the error for the reconstruction of the true state conditioned on the observations, we can strive to obtain a latent representation of the sparse observations consistent to that of the full states. 
This approach is similar to the Generalized Latent Assimilation (GLA) representation proposed by \citet{cheng2022generalised}, but with the addition of matching the means and variances of the state and observation encoders, as illustrated in Figure \ref{fig:vae_obsmatching}. Experimentally, training the encoders without matching the latent representations can lead to a solution where the latent observations are segregated from the latent states, yet result in sufficient reconstructions. Using these two separate encoders, we can match the latent representations such that we are able to apply EnSF in a full-dimensional manner, even with extremely sparse observations.

Some early papers have utilized autoencoder architectures to conduct data assimilation in various ways. GLA proposed in \citet{cheng2022generalised} 
uses the EnKF to assimilate observations in the latent space of the VAE. However, for latent spaces with high dimensions, EnKFs are computationally expensive and also subject to the curse of dimensionality. Meanwhile, overly constricting the latent dimension to simplify the computation would lead to magnified reconstruction errors, so the GLA algorithm needs to strike a balance between efficiency and accuracy. Additionally, depending on the type of architecture, local structures may be lost when encoding the state vector into the latent vector, so localized methods such as the LETKF \citep{HUNT2007112} can lose their advantage.

\subsection{Latent ensemble score filter}
For simplicity, let the latent state $\xi_t$ and latent data $\zeta_t$ be defined as the encoded state and data as 
\begin{equation}
    \xi_t = (\mu_t, \sigma_t^2) = \mathcal{E}(x_{t}) \text{ and } \zeta_t = (\mu_t^{\text{obs}}, (\sigma_t^{\text{obs}})^2) = \mathcal{E}_{\text{obs}}(y_{t}),
\end{equation}
respectively. We conduct data assimilation in the latent space by assimilating the latent data $\zeta_t$ to the latent state $\xi_t$. Mathematically, the latent state $\xi_t$ follows approximately (up to the VAE reconstruction error) the latent dynamical system  
\begin{equation}\label{eq:transition-latent}
    \xi_t \approx \mathcal{E}(M(\mathcal{D}(z_{t-1}(\xi_{t-1})), \varepsilon_t))
\end{equation}
as a result of the full dynamical system in \Eqref{eq:transition} and the VAE with the reparametrization $z_{t-1}(\xi_{t-1}) = \mu_{t-1} + \sigma_{t-1}\cdot \epsilon_{t-1}$ at time $t-1$. The latent data $\zeta_t$ approximately satisfies
\begin{equation}\label{eq:observation-latent}
    \zeta_t = \mathcal{E}_{\text{obs}}(H(\mathcal{D}(z_{t}(\xi_{t})) + \gamma_t)
\end{equation}
as a result of the observation map in \Eqref{eq:observation} and the VAE with the  reparametrization $z_{t}(\xi_{t}) = \mu_{t} + \sigma_{t} \cdot \epsilon_{t}$ at time $t$. When calculating the likelihood of the observation in the latent space, we can simply \Eqref{eq:observation-latent} and use an additive latent observation noise $\hat{\gamma}_t$ estimated from the full space observation noise $\gamma_t$ through the observation encoder. As the coupled encoders seek to match the full states and the sparse observations in their latent representations, we can further approximate the latent observation map in \Eqref{eq:observation-latent} as an identity map, i.e., $\zeta_t \approx \xi_t + \hat{\gamma}_t$, which simplifies and accelerates the computation of the gradient of the log-likelihood function by avoiding automatic differentiation through the map in \Eqref{eq:observation-latent} with respect to the latent state. 


With the latent state following the latent dynamics, and the approximate latent observations, we consider the Bayesian filtering problem in the latent space by formulating the prediction step as 
\begin{equation} \label{eq:prediction_ensf}
\textbf{Prediction: } P(\xi_{t} | \zeta_{1:t-1}) = \int P(\xi_{t} | \xi_{t-1}) P(\xi_{t-1} | \zeta_{1:t-1}) d\xi_{t-1},
\end{equation}
with the transition probability $P(\xi_{t} | \xi_{t-1})$ detemined by the latent dynamical system in \Eqref{eq:transition-latent}. The update step for the posterior of the latent state $\xi_t$ given the latent data $\zeta_t$ is defined as
\begin{equation} \label{eq:bayes_filtering_ensf}
\textbf{Update: } P(\xi_{t} | \zeta_{1:t}) = \frac{P(\zeta_{t} | \xi_{t}) P(\xi_{t} | \zeta_{1:t-1})}{P(\zeta_{t} | \zeta_{1:t-1})},
\end{equation}
where $P(\zeta_{t} | \xi_{t})$ is the likelihood of the latent data given latent state $\xi_t$, which can be computed by the latent observation map in \Eqref{eq:observation-latent}. The latent normalization term $P(\zeta_{t} | \zeta_{1:t-1}) = \int P(\zeta_{t} | \xi_{t}) P(\xi_{t} | \zeta_{1:t-1}) d \xi_{t}$ is also generally intractable. 

We employ the diffusion-based ensemble score filter in Section \ref{sec:DMEnSF} for the above Bayesian filtering problem in the latent space by replacing the state $x_t$ with the latent state $\xi_t$, and the observation $y_t$ with the latent observation $\zeta_t$. Note that the latent state $\xi_t$ and the latent observation $\zeta_t$ have the same dimension $2r \ll d$. We present one step of the Latent-EnSF in Algorithm \ref{alg:ensf-latent}. 






\begin{algorithm}
\caption{One Step of Latent EnSF}\label{alg:ensf-latent}
\begin{algorithmic}
\Require Ensemble of the states $\{x_{t-1}\}$ from distribution $P(x_{t-1} | y_{1:t-1})$ and the observation $y_{t}$. State encoder $\mathcal{E}$, observation encoder $\mathcal{E}_{\text{obs}}$, and decoder $\mathcal{D}$.
\Ensure Ensemble of the states $\{x_{t}\}$ from the posterior distribution $P(x_{t} | y_{1:t})$.
\State Encode the state $\xi_{t-1} = \mathcal{E}(x_{t-1})$ and the observation $\zeta_t = \mathcal{E}_{\text{obs}}(y_t)$. 
\State Run the (stochastic) dynamical system model in \Eqref{eq:transition-latent} from $\{\xi_{t-1}\}$ to obtain samples $\{\xi_t\}$. 
\For{$\tau = \tau_k,...,\tau_0$}
\State Estimate the prior score $\nabla_\xi \log P(\xi_{t, \tau} | \zeta_{1:t-1})$ using \Eqref{eq:prior_score} and \ref{eq:weight} in the latent space.
\State Estimate the posterior score $\nabla_\xi \log P(\xi_{t, \tau} | \zeta_{1:t})$ using \Eqref{eq:scorefunction} in the latent space.
\State Solve the reverse-time SDE in \Eqref{eq:reverse_sde} to generate the ensemble of latent states $\{\xi_t\}$. 
\EndFor
\State Sample $z_t = \mu_t + \sigma_t \cdot \epsilon_t$ from the ensemble of the latent states $\xi_t = (\mu_t, \sigma_t)$ and samples of $\epsilon_t$. 
\State Decode the ensemble of latent variables $\{z_t\}$ to the ensemble of the full states by $x_t = \mathcal{D}(z_t)$. 
\end{algorithmic}
\end{algorithm}


Because the sparse observations have been mapped to the same dimension as the state in the latent space, we are able to circumvent the weakness of the EnSF with vanishing gradient of the log-likelihood function when dealing with the sparse observation problems. In addition, since the encoder is a flexible neural-network representation, the model is able to restrict the latent observation map $H_{\text{latent}}$
in the latent space to be the identity function $H_{\text{latent}}(\xi_t) = \xi_t$, allowing for analytical solutions for the score function without the need for automatic differentiation. This means that the model can leverage the full computational speed-up of the EnSF in the latent space. Finally, the regularization done by the VAE such that the latent vector tends to follow a $N(0, 1)$ distribution simplifies hyperparameter search of the appropriate noise $\gamma_t$ for real world problems.

\subsection{Addressing EnSF's numerical instability on small scales}
In practice, when applying the EnSF to problems where the values of the states and their variances are very small, specifying the corresponding observation noise such that $\gamma_t \ll 0.05$ can result in numerical instability issues when sampling the posterior distribution by reverse time SDE with the drift and diffusion terms given in \Eqref{eq:fg}.  

This numerical instability issue is also present in the VAE latent space where the components are regularized such that $\mu_t \approx 0$ and $\sigma_t \approx 1$. We address this issue by multiplying the latent representations with a scalar constant $\psi_{\text{latent}}$, conducting the EnSF for the rescaled latent representation, and subsequently scaling the posterior samples back to the original latent representation. We conducted a small grid search experiment when working with the shallow water wave propogation problem in Section \ref{section:shallow_water}, where $\psi_{\text{latent}} = 500$ seemed to work reasonably.

Remarkably, this $\psi_{\text{latent}}$ applies well in all our experiments, which implies that this scaling hyperparameter stays fairly steady across all problems. Intuitively, this is consistent; the latent regularization conducted by the VAE constrains the latent space to be in a similar range. In contrast, when EnSF is conducted on the full state space, such advantages are lost and a different scaling parameter $\psi$ is needed for different problems with different scales of the state and observation noise.

\section{Experiments}
\label{sec:experiments}

In this section, we illustrate the limitation of using the EnSF for sparse observations, and demonstrate the improved accuracy and fast convergence of the Latent-EnSF compared to several data assimilation methods for a synthetic complex physical system modeled by shallow water equations with sparse observations in both space and time. We also demonstrate the merits of Latent-EnSF for medium-range weather forecasting with real-world data and a ML-based dynamical system. 

\subsection{Shallow water wave propagation} \label{section:shallow_water}

We consider the propagation of shallow water waves described by shallow-water equations \citep{vreugdenhil1994numerical}, which are a system of hyperbolic PDEs that describe fluid flow of free surface with the conservation of momentum and mass, with three state variables including the water height and two components of the water velocity field. Such models find practical applications in hydrology for predicting flood waves and in oceanography for modeling tsunami wave propagation. 

In the experiment, we consider water wave propagation with the initial water displacement modeled by a local Gaussian bump perturbation from the flat surface in a square domain of size $L \times L$ with $L=10^{6}$ m in each direction and a constant depth $h = 100$ m for the topography of the floor. For the simulation, we use a finite difference code adapted from \citet{jostbr}, with the domain discretized into a uniform grid of $150 \times 150$. 
The simulation is then run for 2000 time steps using a upwind scheme with a time step $\Delta t = 0.1 \frac{\Delta x}{\sqrt{gh}}$ with $\Delta x = L/150$ to satisfy the Courant–Friedrichs–Lewy condition \citep{cfl}, see Figure \ref{fig:samples_grid} for the evolution of the height and its sparse observations with 4x ($75 \times 75$) and 225x ($10 \times 10$) data reduction.

For the data assimilation, we start with a mis-estimated initial state where the Gaussian bump perturbation is shifted away from the true initial state. 100 ensemble members are then taken by adding a negligible amount of noise to the mis-estimated state. The data assimilation is conducted with observations every 20 time steps unless otherwise stated, for a total of 100 assimilation steps, with an observation noise variance of 1. When conducting the EnSF, we take 100 forward Euler steps to solve the reverse-time SDE, and use hyperparameters $\epsilon_\alpha = 0.05$, $\epsilon_\beta = 0$.



\subsubsection{EnSF and Sparsity} \label{section:ensf_sparsity_experiments}

In Section \ref{section:ensf_sparsity}, we illustrated the vanishing gradient of the log-likelihood function for the dimensions in which observations are not available. Here we empirically demonstrate this effect on EnSF for data assimilation with sparse observations at different levels of sparsity.



\begin{figure}[h]
\begin{minipage}[c]{0.535\linewidth}
    \includegraphics[width=\textwidth]{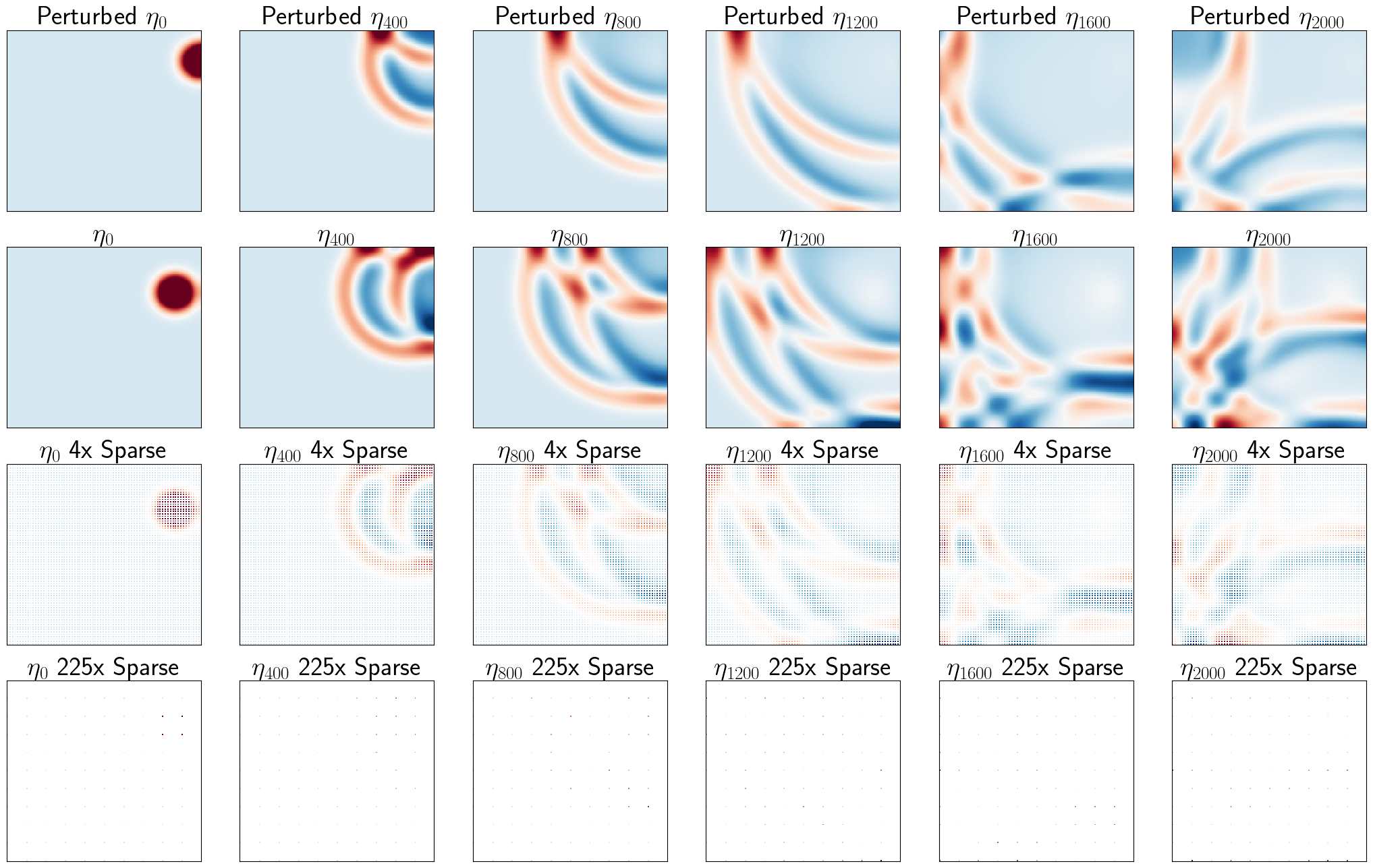}
    \caption{Evolution of states and sparse observations.}
    \label{fig:samples_grid}
\end{minipage}%
\hfill
\centering
\begin{minipage}[c]{0.415\linewidth}
\includegraphics[width=\textwidth]{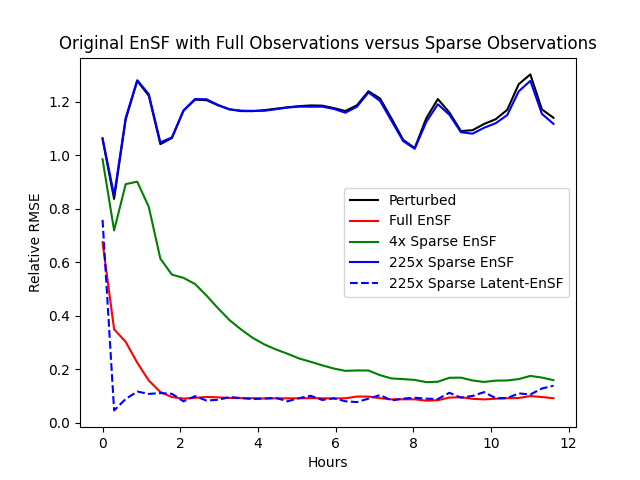}
    \caption{Relative RMSE of EnSF and Latent-EnSF for sparse observations.}
    \label{fig:ensf_sparsity}
\end{minipage}
    
\end{figure}

In Figure \ref{fig:ensf_sparsity}, we plot the performance in terms of relative root mean square error (RMSE) of the assimilated state as the observation space becomes increasingly sparse. The relative error of the perturbed trajectory and the original trajectory is also plotted as a reference. We can observe that when using the EnSF, increasing the sparsity of the observation data leads to increasing assimilation errors and slower convergence to the true state. In particular, for extreme cases such as when the observation data are reduced by a factor of 225 times, there is almost no discernible difference between the assimilated error of EnSF compared to a non-assimilated trajectory, or EnSF does not bring any improvement in the data assimilation. In contrast, the Latent-EnSF achieves small relative errors with fast convergence to the true state even in the case of extreme sparsity, with the errors at the same level achieved by EnSF with the observation of the full state. These findings align with \citet{bao2024nonlinearensemblefilteringdiffusion} where the sparsity in their experiments for EnSF is limited to a 2x reduction.

\subsubsection{Latent-EnSF Results}

\begin{figure}
\begin{subfigure}[h]{0.48\linewidth}
\includegraphics[width=\textwidth]{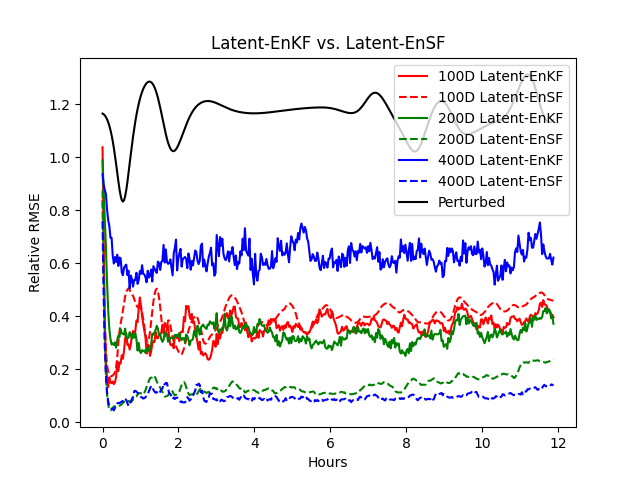}
\end{subfigure}
\hfill
\begin{subfigure}[h]{0.48\linewidth}
\includegraphics[width=\textwidth]{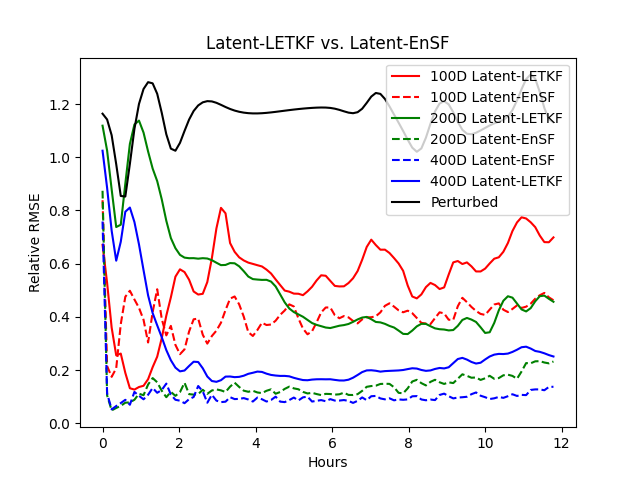}
\end{subfigure}%
\caption{Relative RMSE of Latent-EnSF compared to Latent-EnKF (left) and Latent-LETKF (right) with varying latent dimensions, with the error of the perturbed state shown as a reference. 
}
\label{fig:ensf_latent_dims}
\end{figure}

We train the coupled VAE as shown in Figure \ref{fig:vae_obsmatching} using convolutional layers with input dimension $3 \times 150\times 150$ and latent dimension $4 \times 10 \times 10 = 400$ with 4 channels for the experiments other than that in Figure \ref{fig:ensf_latent_dims}, which is for the evaluation of the performance with respect to the latent dimension size. The following experiments use a $10 \times 10$ observation grid, corresponding to the 225x sparse entry in Figure \ref{fig:ensf_sparsity}. We compare the Latent-EnSF and the Latent-EnKF, which shares the same VAE weights as the Latent-EnSF. For the approximate $\gamma_t$ observation noise terms in the latent space, we adopt a heuristic by taking the mean standard deviation values given by encoded latent states for all three approaches. In Figure \ref{fig:ensf_latent_dims}, we can see that the Latent-EnKF struggles with higher dimensional latent spaces, whereas the Latent-EnSF gets more accurate as a result of a decrease in reconstruction error for the VAE, even though they perform similarly for the 100-dimensional case. 

In addition to the Latent-EnKF, we test against the Latent-LETKF in high-dimensional latent space, as shown in the right part of Figure \ref{fig:ensf_latent_dims}. This utilizes the LETKF which in practice handles high-dimensional data assimilation much better than the EnKF. However, we can see that though it indeed beats the Latent-EnKF benchmark, especially with higher latent dimensions, our approach is still much better in terms of assimilation speed, accuracy, and efficiency (see Table \ref{tab:speed}) for all cases.

\begin{figure}[!htb]
    \centering
    \includegraphics[width=1\linewidth]{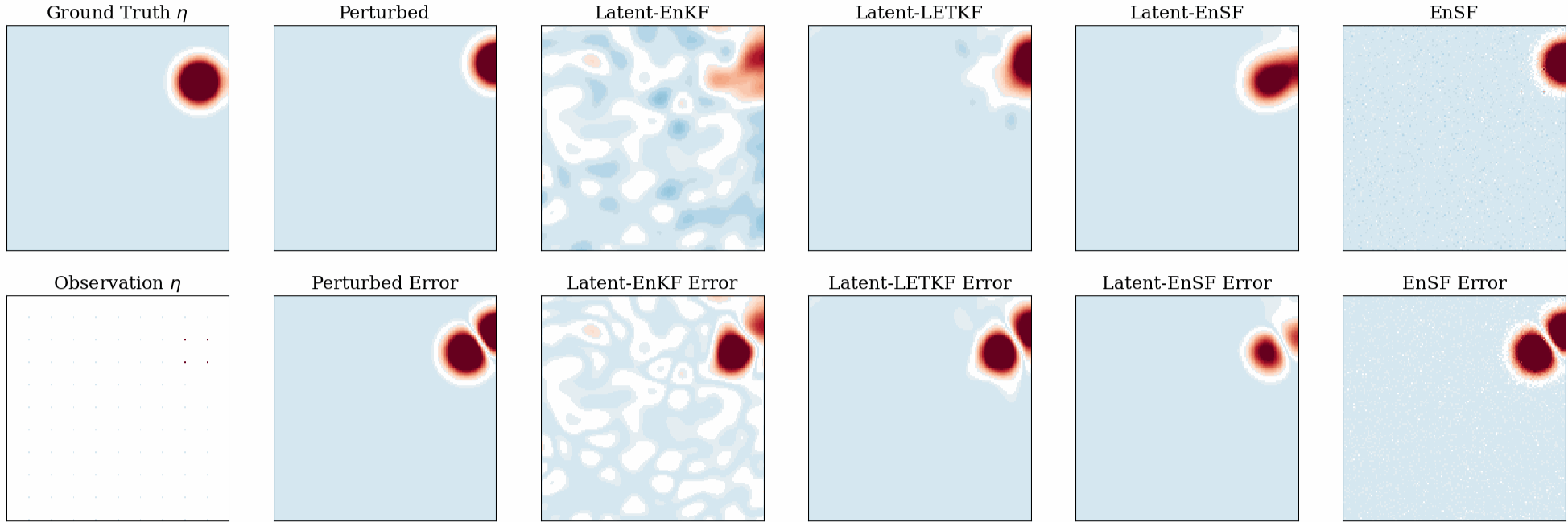}    
    \includegraphics[width=1\linewidth]{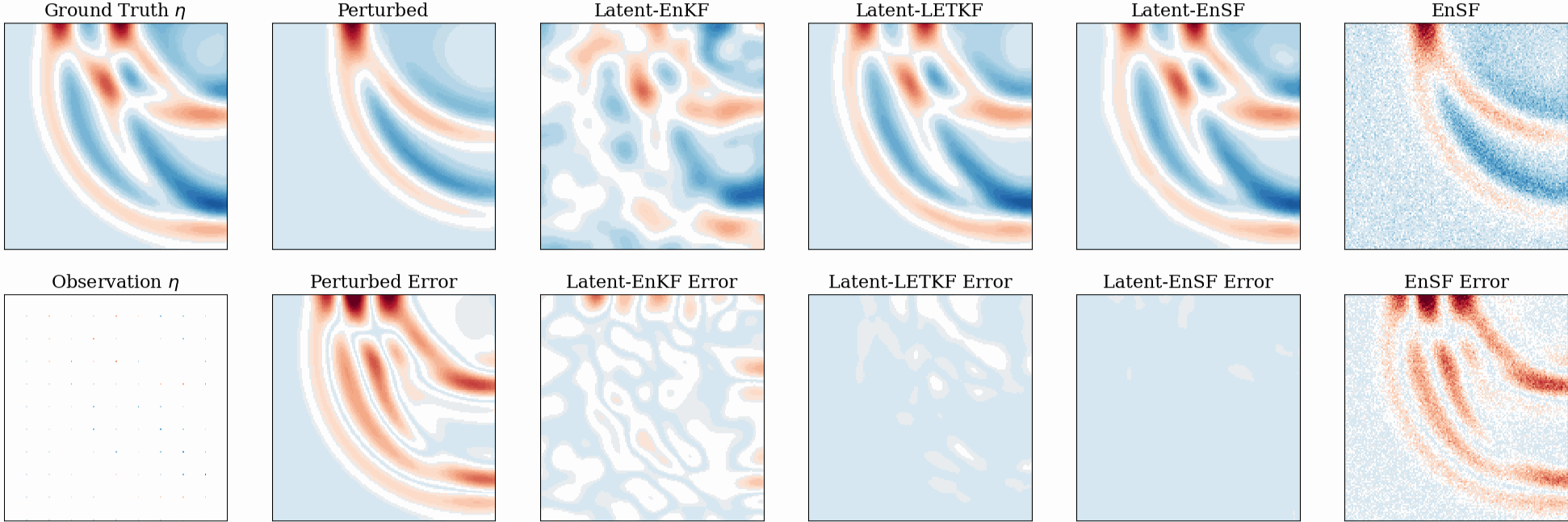}
    \includegraphics[width=\textwidth]{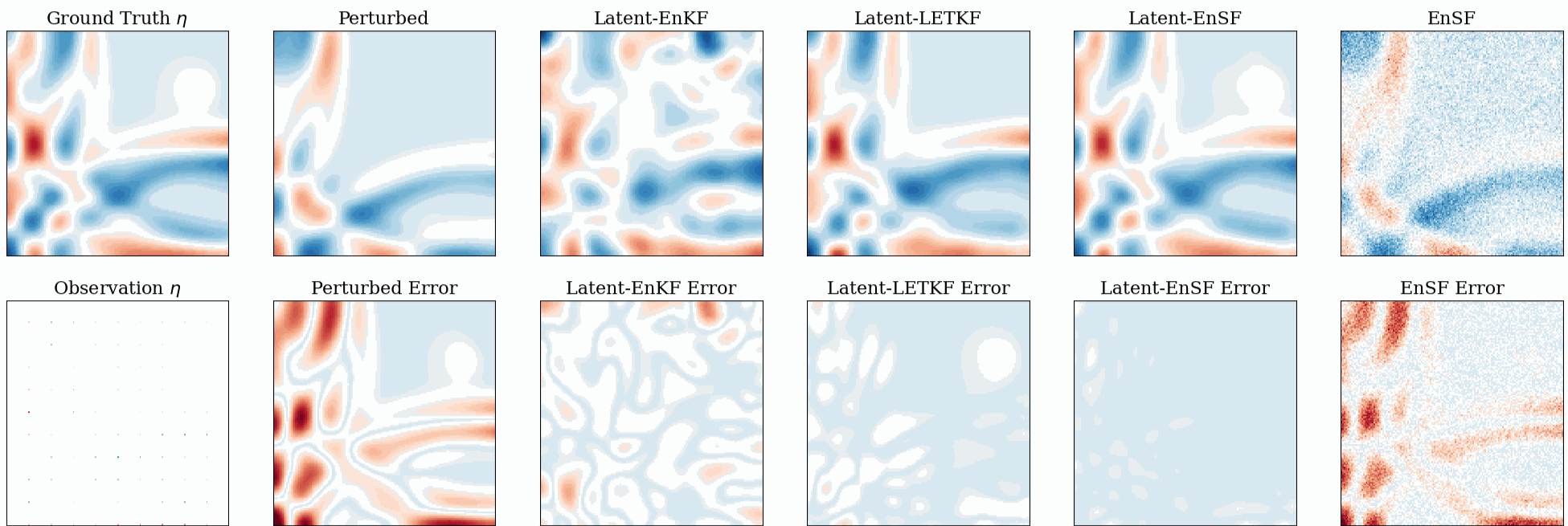}
    \caption{Comparison of different methods at  time step 1 (top two rows), 1000 (middle two rows), and 2000 (bottom two rows), with the ground truth, perturbed states, and assimilated states on the top and their corresponding errors from the ground truth at the bottom of each two rows.
    }
    \label{fig:swe_sample}
\end{figure}

Figure \ref{fig:swe_sample} shows the comparison of the assimilated results of sparse observations using different methods at the first time step, time step 1000, and 2000 (the final time step). We can observe that even though the perturbed dynamics is very far from the ground truth, the Latent-EnSF is able to push the perturbed dynamics to match the ground truth very well with small error, while EnSF fails to do so with its error very close to the difference between the perturbed and true states. Moreover, Latent-EnSF performs the best among the data assimilation methods in the same latent space.

\begin{figure}[h]
    \centering
    \includegraphics[width=0.5\textwidth]{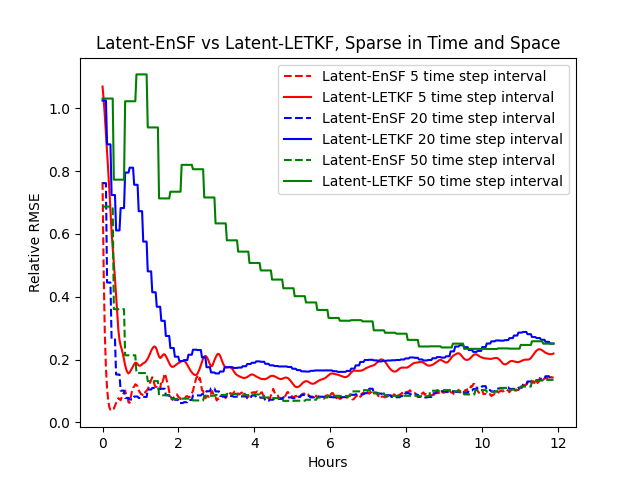}
    \caption{Relative RMSE of Latent-EnSF vs Latent-LETKF for varying observation sparsity in time, with the assimilation frequency at every 5, 20, and 50 time steps, respectively.  
    }
    \label{fig:ensf_sparse_time}
\end{figure}

In Figure \ref{fig:ensf_sparse_time}, we report the comparison of Latent-EnSF and Latent-LETKF for their performances with different temporal sparsity in the observations, which come in every 5, 20, and 50 time steps with increasing sparsity. As we can observe, the assimilation accuracy of Latent-EnSF is not affected by the increased sparsity; only the burn-in time is slightly increased. In contrast, the Latent-LETKF is very sensitive to the temporal sparsity, with much slower convergence for larger sparsity, while Latent-EnSF is robust to sparse observations not only in space but also in time in this experiment.

\subsection{Medium-Range Weather Forecasting}

Sparse data assimilation is particularly important for accurate medium-range weather forecasting due to the complexity of the problem which can skew results of machine learning models. The commonly used dataset for conducting these weather forecasting experiments is the ERA5 \citep{era5}, coming from the European Centre for Medium-Range Weather Forecasts (EMCWF). The main variables at play in this dataset include the wind speeds, temperature, geopotential, and humidity, just to name a few, all of which are available at any of the 137 different pressure levels. This dataset comes with 40 years of reanalyzed data, and can support various degrees of fidelity. Currently, ML-based prediction approaches such as FourCastNet \citep{pathak2022fourcastnetglobaldatadrivenhighresolution}, Pangu Weather \citep{bi2022panguweather3dhighresolutionmodel}, and GraphCast \citep{lam2023graphcastlearningskillfulmediumrange} have shown to be extremely efficient when applied to weather forecasting compared to standard PDE-based approaches. We adopt the 21 variables used in the FourCastNet paper which include wind velocities and geopotential at 3 different pressure levels, along with relative humidity and temperature at a few key pressure levels and the total column water vapor at the surface. For forecasting, we train a FourCastNet model on our coarse subset of the dataset. When training the VAE for data assimilation, we create a ResNet-style architecture similar to what we constructed in the shallow-water experiments. The encoder downsamples by a factor of 2 twice between convolution layers, resulting in a latent dimension size of $32 \times 36 \times 18 = 20736$. In between the layers we use Leaky-ReLU layers with a coefficient of 0.2. We use this as a demonstrative task to show the feasibility of applying Latent-EnSF to complex systems, so we train and test on 10 years of data with $2.5^{\circ}$ resolution, with a grid of size $144 \times 72$, as a proof of concept. Evaluations are done for the year 2015, where we average the metrics across nine 41-day windows. Data assimilation is done once a day, incorporating observations at a 64x sparsity rate, with 8x reduction in each dimension, or a grid of $18 \times 9$ observations.

\begin{figure}[h]
\begin{minipage}[c]{0.48\linewidth}
\includegraphics[width=\textwidth]{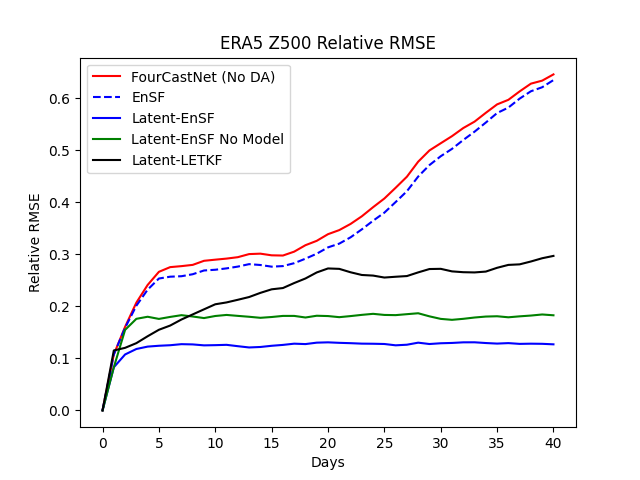}
\label{fig:era5_relative_rmse_z500}
\end{minipage}
\hfill
\begin{minipage}[c]{0.48\linewidth}
\includegraphics[width=\textwidth]{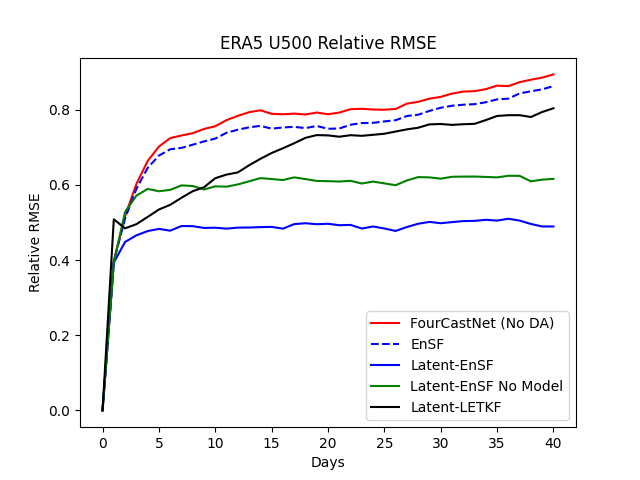}
\label{fig:era5_relative_rmse_u500}
\end{minipage}%
\centering
\caption{Relative RMSE for the assimilation of Z500 (geopotential at 500hPA) and U500 (u-component of wind at 500hPA) on the ERA5 with respect to the number of days of forecasting.}
\label{fig:era5_rmse}
\end{figure}

As shown in Figure \ref{fig:era5_rmse}, the forecasting by the FourCastNet-based ML model without data assimilation becomes increasingly inaccurate because of the accumulated errors, which limits its relatively reliable prediction for the first 14 days. The data assimilation errors by EnSF follow closely those of FourCastNet forecasting without data assimilation, which implies that EnSF fails with the data assimilation with sparse observations. 
In contrast, the Latent-EnSF approach quickly levels off the error in the first couple of days and preserves the accuracy for all the following days, despite the increasing inaccuracy of the FourCastNet forecasting. Note that the Latent-LETKF leads to increasing assimilation errors, much higher than Latent-EnSF, with increasingly inaccurate FourCastNet forecasting. To demonstrate that the FourCastNet is helpful in the data assimilation, we also run the Latent-EnSF without a known dynamical model, in which case we assume the dynamics of the state does not evolve, i.e., $x_{t+1} = M(x_{t}) = x_t$. From the comparison of Latent-EnSF No Model and Latent-EnSF (with FourCastNet), we observe that the ML-based FourCastNet forecasting does help for the data assimilation, even though it becomes increasingly inaccurate in the long run.

\begin{figure}
    \centering
    \includegraphics[width=1\linewidth]{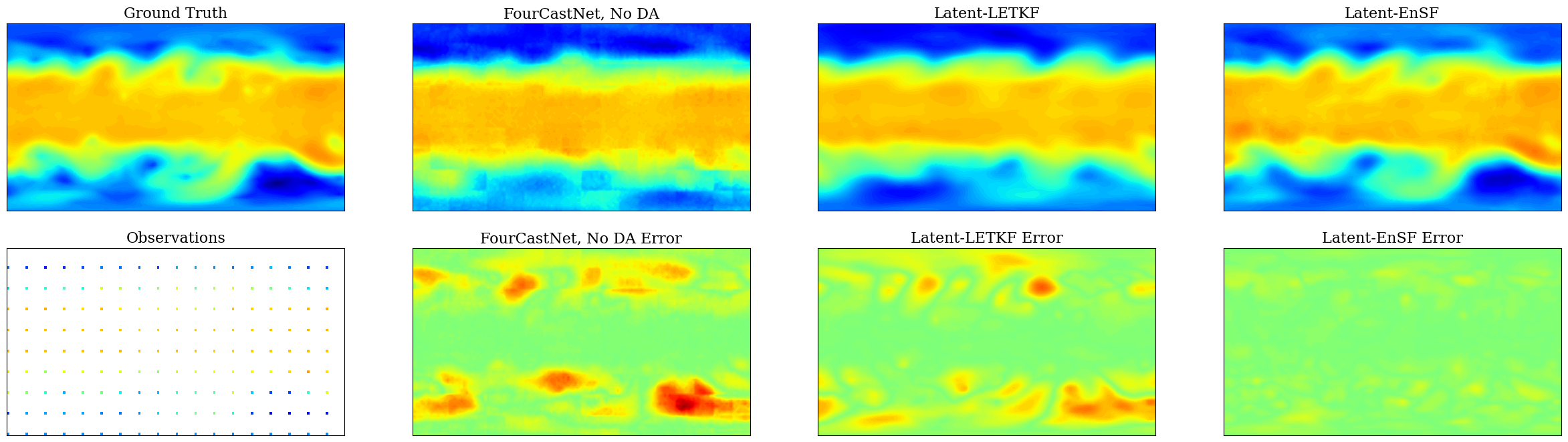}
    \includegraphics[width=1\linewidth]{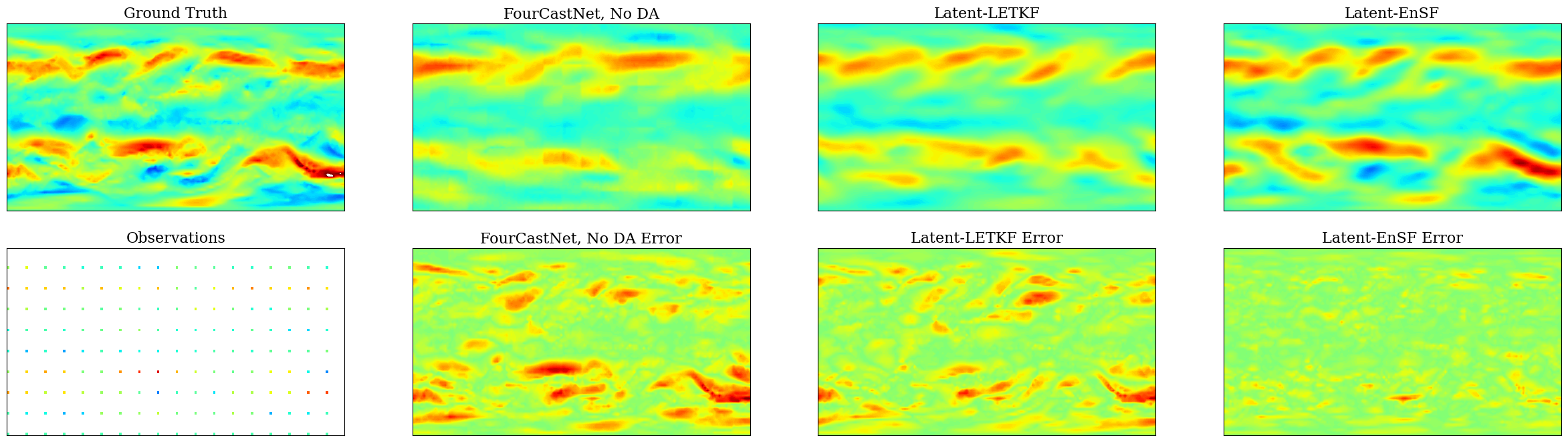}
    \caption{ERA5 Z500 (top two rows) and U500 (bottom two rows) medium-ranged weather forecasting samples after 41 days, along with the errors. Data assimilation is conducted once a day with 64x sparse observations. We compare against a baseline FourCastNet model and Latent-LETKF.}
    \label{fig:era5-z500}
\end{figure}

We observe in Figure \ref{fig:era5_rmse} that the assimilation for the state Z500 is much more accurate than that of U500, which can also be observed in Figure \ref{fig:era5-z500} after 41 days of forecasting by FourCastNet and by data assimilation. This is due to the much higher state complexity with rich local characteristics and sharp changes in the u-component of the wind velocity U500, which would demand more data (with less sparsity) beyond the very sparse $18\times 9$ observations for more accurate data assimilation. 

To examine the computational efficiency, we report the mean and standard deviation of the time it takes the same machine (a combination of an Nvidia A6000 GPU with an AMD 7543 CPU) for one step of the assimilation by each of the methods in Table \ref{tab:speed} for the two applications. The latent dimension $2r$ is applicable to all methods except EnSF with dimension $d$ in the full space. 
We can see that even though the Latent-LETKF leads to higher accuracy than Latent-EnKF, it takes much more time than all the other methods, due to the need to iterate through the dimensions to construct each individual localization matrix (this could be ameliorated by parallel computing in many processors). However, the computational cost increases approximately linearly with respect to the dimensions compared to the much faster increase of the Latent-EnKF. In comparison, the Latent-EnSF achieves the highest accuracy, fastest convergence, and still uses the smallest time. By comparison of the two applications, we see that although the latent dimensions differ by a scale of 50, the Latent-EnKF takes over 700 times as long for assimilation in the higher latent dimension, compared to 68 times for Latent-LETKF and only 4 times for Latent-EnSF, which implies the much better scalability of the Latent-EnSF compared to the other methods.
Additionally, note that even with the added complexity of doing a forward pass through the encoder and decoder, the assimilation by Latent-EnSF is still faster than that by EnSF in the full space, making Latent-EnSF highly conducive for real-time data assimilation.

\begin{table}[h]
\caption{Mean and standard deviation of wall-clock time in seconds of different assimilation approaches for both our experiments, obtained using Nvidia A6000 GPU + AMD 7543 CPU.}
\label{tab:speed}
\begin{tabular}{ |c|c|c|c|c|c|} 
\hline
Dataset & Latent Dim & Latent-EnKF & Latent-LETKF & Latent-EnSF & EnSF \\
\hline
SWE& 400 & $0.056 \pm 0.002$ & $6.567 \pm 0.069$ & $0.051 \pm 0.001$ &  $0.223 \pm 0.002$\\
ERA5 & 20736 & $41.161 \pm 0.640$ & $450.213 \pm 3.512$ & $0.198 \pm 0.008$ & $0.501 \pm 0.001$\\
\hline
\end{tabular}
\end{table}

\section{Conclusion}

In this paper, we introduced Latent-EnSF, a novel data assimilation method in addressing the joint challenges of high dimensionality in the state and high sparsity in the observation for nonlinear Bayesian filtering problems, which have been the key and common challenges for the existing approaches. We proposed a coupled VAE for information compression to the latent space, which is trained for a consistent match of the full states and sparse observations for their latent representations and with latent regularization, as well as for a consistent reconstruction of the full states. We also presented the corresponding approximate latent dynamical system and the simplified latent observation map in accelerating the sampling process in the latent space. We demonstrated the high accuracy, fast convergence, and high efficiency of the proposed method compared to other approaches for a challenging data assimilation problem with PDE-based shallow water wave propagation model and synthetic data that is sparse in both space and time. We also applied and demonstrated the advantage of the proposed method to a ML-based model with real-world data for data assimilation in medium-range weather forecasting. 



For further avenues of interest, extending Latent-EnSF to incorporate latent dynamics could further enhance the computational efficiency. Additionally, exploring more complex architectures could allow us to conduct data assimilation for unstructured observation points and continuous-time data where extremely sparse observations arrive in a stochastic manner. Overall, Latent-EnSF presents a promising direction for advancing data assimilation techniques and bringing an efficient way of utilizing machine learning models to simulate real-world systems with improved accuracy.

\section*{Acknowledgement} This work is partially supported by NSF grant \# 2325631, \# 2245111, and \# 2245674. We acknowledge helpful discussions with Prof. Felix Herrmann, Dr. Jinwoo Go, and Pengpeng Xiao.

\bibliographystyle{iclr2024/iclr2024_conference.bst}
\bibliography{references}







\end{document}